\documentclass[letterpaper, 10 pt, conference]{ieeeconf}  
\IEEEoverridecommandlockouts                                      
\usepackage{xcolor}
\long\def\invis#1{}

\newtheorem{theorem}{Theorem}[section]
\newtheorem{problem}[theorem]{Problem}

\usepackage{algpseudocode}
\usepackage{algorithm}
\usepackage{graphicx}
\usepackage{cite}

\usepackage{xcolor}

\title{\LARGE \bf
\textsc{AG-Cvg}: Coverage Planning with a Mobile Recharging UGV \\and an Energy-Constrained UAV
}

\author{Nare Karapetyan$^{1}$, Ahmad Bilal Asghar$^{2}$, Amisha Bhaskar$^{2}$, Guangyao Shi$^{3}$, \\Dinesh Manocha$^{2}$ and Pratap Tokekar$^{2}$
\thanks{ This work is supported in part by National Science Foundation Grant No. 1943368 and Army Grant No. W911NF2120076.}
\thanks{$^{1}$Woods Hole Oceanographic Institution (WHOI), Woods Hole, MA 02543, USA. Emails: {\tt\small nare@whoi.edu}}
\thanks{$^{2}$Maryland Robotics Center (MRC), University of Maryland, College Park, MD 20742, USA. Emails: {\tt\small \{abasghar, amishab, gyshi, dmanocha, tokekar\}@umd.edu}}%
\thanks{$^{3}$University of Southern California, Los Angeles, CA 90007, USA. Emails: {\tt\small shig@usc.edu}}
}

\begin{document}

\maketitle
\thispagestyle{empty}
\pagestyle{empty}


\begin{abstract}

In this paper, we present an approach for coverage path planning for a team of an energy-constrained Unmanned Aerial Vehicle (UAV) and an Unmanned Ground Vehicle (UGV). Both the UAV and the UGV have predefined areas that they have to cover. The goal is to perform complete coverage by both robots while minimizing the coverage time. The UGV can also serve as a mobile recharging station. The UAV and UGV need to occasionally rendezvous for recharging. We propose a heuristic method to address this NP-Hard planning problem. Our approach involves initially determining coverage paths without factoring in energy constraints. Subsequently, we cluster segments of these paths and employ graph matching to assign UAV clusters to UGV clusters for efficient recharging management. We perform numerical analysis on real-world coverage applications and show that compared with a greedy approach our method reduces rendezvous overhead on average by 11.33$\%$. We demonstrate proof-of-concept with a team of a VOXL m500 drone and a Clearpath Jackal ground vehicle, providing a complete system from the offline algorithm to the field execution.
\end{abstract}

\section{INTRODUCTION}
\label{section:introduction}

The area coverage problem appears prominently in various domains where robots are used as sensors including agriculture~\cite{santos2020path}, painting~\cite{zhou2022building}, and environmental monitoring~\cite{dunbabin2012robots,karapetyan2021meander,sung2023decision}. Many of these applications can benefit from using a team of heterogeneous robots such as a team of UAV and UGV or UAV and an Unmanned Surface Vehicle (USV). One of the bottlenecks of UAVs for long-term operation is their short battery life. Using UGVs that are capable of acting as recharging stations for the UAVs can both expand the regions to be visited and extend the duration of the execution~\cite{tokekar2016sensor,yu2019algorithms,yu2019coverage}.


While an extensive body of work has been done to address the area coverage problem \cite{galceran2013survey,karapetyan2021robot, Acar2002J2, lumelsky1989sensor, singh2008mobile, hollinger2013sampling}, the simultaneous coverage path planning problem with a team of UGV and energy-constrained UAV has not been addressed in the literature. When the energy constraint is considered,  we need to balance potentially conflicting objectives: coverage performances of the UAV and the UGV, and guarantee feasible recharging rendezvous. Moreover, the complexity of the problem will increase, since representing the energy states requires an additional dimension in the search space~\cite{shi2022risk}. 

\begin{figure}[t]
    \centering
    {\includegraphics[width=1\columnwidth]{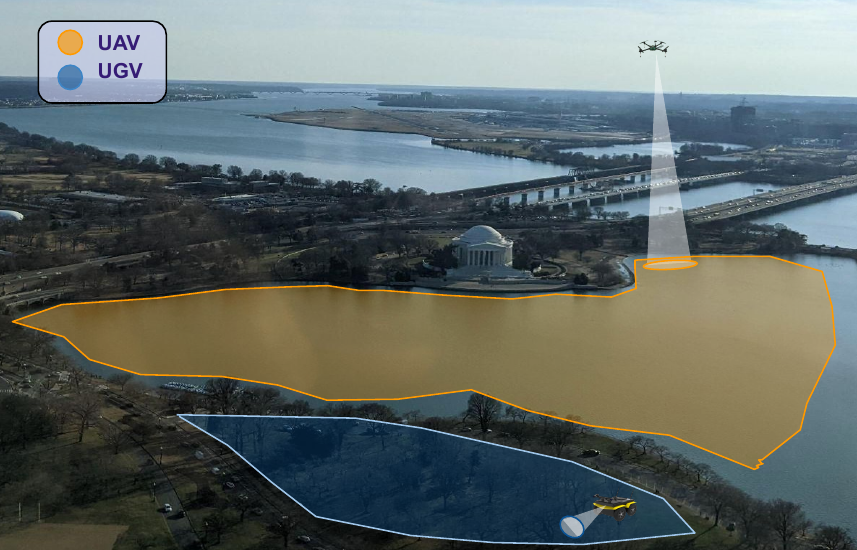}}
    \caption{Sample allocation of task regions: UGV surveying a large terrain with obstacles and UAV a lake surface.}
    \label{fig:beauty_shot}
\end{figure}

The most relevant work by Yu et al.~\cite{yu2019coverage} solves the coverage problem for an energy-constrained UAV while a UGV serves as a recharging station. The problem is formulated as a Generalized Travelling Salesman Problem to minimize the coverage time of the UAV along with takeoff and landing time. However, in this work, the UGV serves only as a recharging agent and performs no coverage. 

In this paper, we address the coverage problem for both the energy-constrained UAV and the UGV that will also act as a mobile recharging station. In this formulation, we assume that each of the robots has its own assigned region to cover (Fig.~\ref{fig:beauty_shot}). We design a complete coverage pipeline that takes the area to be covered as input and produces paths for UAV and UGV that not only provide coverage but also respect UAV's energy constraint by finding appropriate rendezvous locations. We propose the Air-Ground Coverage Path Planning Algorithm (\textit{\textsc{AG-Cvg}}), which first builds coverage paths without taking energy constraints into account. Following this, it segments these paths and utilizes graph matching techniques to allocate UAV path clusters to UGV path clusters, reducing distances traveled for rendezvous on average by $11.33\%$, with up to 25$\%$ improvement on some instances.

\begin{figure*}[t!]
    \centering
    {\includegraphics[width=1.95\columnwidth]{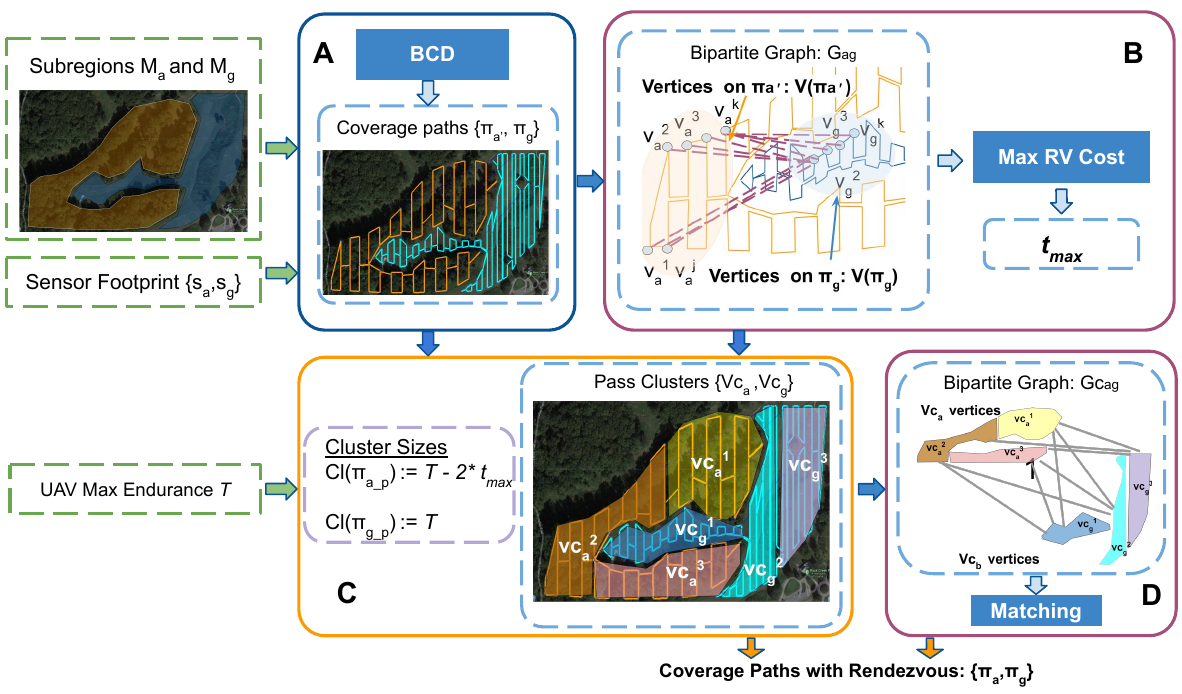}}
    \caption{Complete Pipeline of \textsc{AG-Cvg} algorithm (see Algorithm~\ref{algorithm:ag_cover}). It consists of four main steps: (A) finding the coverage paths; (B) finding the worst-case rendezvous cost; (C) clustering paths into regions that can be covered completely with a UAV without running out of charge; (D) finding the rendezvous locations.}
    \label{fig:pipeline}
\end{figure*}

The contributions of this paper are:
\begin{itemize}
    \item The first coverage path planning algorithm when both UAV and UGV simultaneously perform coverage and the UGV acts as a mobile recharging station.
    \item Feasible recharging of the energy-constrained UAV. 
    \item Real-world deployment system for executing generated trajectories in the field with a VOXL m500 drone and Clearpath Jackal ground vehicle.
\end{itemize}

The rest of this paper is organized as follows. We begin with an overview of the recent works in the field in Section~\ref{section:related_work}. We formally define the problem in Section~\ref{section:problem_formulation} and follow with a description of the proposed \textsc{AG-Cvg} algorithm. Experiments both on numerical results and real-world deployments are discussed in Section~\ref{section:experiments}. Finally, Section~\ref{section:conclusions} summarizes the results and lessons learned.

\section{RELATED WORK}
\label{section:related_work}

Offline area coverage planning is particularly useful for large-scale environments. While learning-based methods can be helpful in incorporating uncertainties about the environment, these approaches are limited to small-scale decision spaces~\cite{lakshmanan2020complete}. Therefore, offline plans can be helpful in breaking down the complexity of large-scale operations, which is especially commonly faced by scientists who perform systematic data collection over extended periods of time. 

The area coverage path planning problem has been extensively studied in robotics~\cite{galceran2013survey, choset2001coverage}. Depending on the goal or constraints of the coverage operation, different variations have been presented such as coverage under limited resources~\cite{strimel2014coverage,agarwal2022area}, complete coverage, information-driven coverage~\cite{paull2012sensor, manjanna2017collaborative,malencia2022adaptive} or sampling-based coverage~\cite{englot2012sampling, hollinger2013sampling}.

The boustrophedon area decomposition approach is a commonly employed technique for achieving complete coverage. It involves dividing the area of interest into cells that are free of obstacles and subsequently directing the robot to traverse each cell in a lawnmower-like pattern to ensure complete coverage~\cite{choset2000coverage}. Many studies have utilized this method as a primary means of addressing coverage problems~\cite{bahnemann2021revisiting}, while others have utilized it as a supplementary technique for improving the quality of coverage~\cite{khan2017online}.

A polynomial-time algorithm was proposed by Xu et al.~\cite{xu2014efficient} for solving single robot coverage that uses a Boustrophedon Cell decomposition (BCD). Here, the problem is represented as a Chinese postman problem (CPP), which ensures an efficient coverage order of cells. The multi-robot extension for vehicles with and without dubin's constraint was proposed as well~\cite{karapetyan2017efficient, karapetyan2018multi}. 

In this paper, we address the energy constraints of the UAV by utilizing the UGV as a mobile recharging station. This cooperative path-planning strategy involving multiple mobile and static recharging stations has been explored in the literature~\cite{liu2022review}.

For instance, this problem has been studied in 1D~\cite{maini2020visibility,maini2018persistent} for persistent visual monitoring of linear features with multi-agent systems that need to be recharged while performing the monitoring mission. The problem has been defined as a Mixed-Integer programming problem and solved through the branch-and-cut approach.
In 2D space, Yu et al.~\cite{yu2019algorithms} discuss the challenges of completing a tour while using a UGV as a station to periodically recharge the UAV. Authors in~\cite{shi2022risk, asghar2022risk} examine the risk-aware version of this problem with stochastic energy consumption. The work presented in~\cite{yu2019coverage} is closely related to our approach, as it also considers mobile recharging for a UAV performing coverage in an environment with limited battery capacity. However, in our study, the UGV has the additional task of covering the environment while providing mobile recharging services to the UAV.

As such to the best of our knowledge the presented work is the first attempt to present a complete system for solving a coverage path planning problem with both a mobile recharging vehicle and an energy-constrained vehicle.

\section{Problem Formulation}
\label{section:problem_formulation}

Consider an environment $M$ and regions of the environment $\{M_g, M_a\}\subseteq M$ representing the areas to be covered by a single UGV and a single UAV respectively. (Note: in the notations, we will use \textit{$a$} subscript to denote UAV variables and \textit{$g$} for UGV variables.) The regions $M_g$ and $M_a$ that may or may not overlap.\invis{Even when regions overlap cooperative coverage might be necessary for different level of awareness.} The UAV has an energy constraint and its maximal endurance in the air while performing coverage is denoted as $T$, including the takeoff time $t_{\mathrm{off}}$ and landing time $t_{\mathrm{lnd}}$. The UGV is equipped with a recharging station and the UAV can land on the UGV to be recharged. We assume that both UAV and UGV maintain constant speeds $\nu_a$ and $\nu_g$ respectively. With this assumption, the maximum distance UAV can travel before discharging is $T*\nu_a$.

 Given the sensor footprints $s_a$ and $s_g$ of the UAV and UGV respectively, the objective is to find paths for each that will provide complete coverage of the regions in a minimal amount of time, while satisfying the UAV's energy constraints. The UAV and UGV must rendezvous for recharging if the size of the mission is larger than the single UAV battery life. 

\invis{
\begin{figure}[h!]
    \centering
    {\includegraphics[width=0.9\columnwidth]{./figures/pass_matching.png}}
    \caption{Cluster Matching phase to find the rendezvous locations. Each of the ${V_a}^i$ and ${V_g}^j$ vertices of Bipartite Graph represents a cluster of coverage passes respectively from UAV's and UGV's coverage path.}
    \label{fig:match}
\end{figure}
}

The problem is formally defined as follows. 
\begin{problem}\label{main_problem}
Given environment $M$ with regions $\{M_g, M_a\}\subseteq M$, sensor footprints $s_a$, $s_g$, and UAV energy constraint $T$, find paths $\pi_a$ and $\pi_g$ in $M$ for the UAV and UGV respectively, such that
\begin{itemize}
    \item The paths $\pi_a$ and $\pi_g$ avoid obstacles.
    \item Each point in the region $M_a$ ($M_g$) is covered by the sensor footprint $s_a$ ($s_g$) along the path $\pi_a$ ($\pi_g$).
    \item The UAV must be able to traverse $\pi_a$ without running out of charge, i.e., in every segment of $P_a$ of length at least $T$, there must be at least one rendezvous with the UGV. 
    \item The total path length $\ell(\pi_a) +\ell(\pi_g)$ is minimized.
\end{itemize}
\end{problem}

Note that the rendezvous not only means that the paths $\pi_a$ and $\pi_g$ must intersect, but also that the UGV and UAV must be at the intersection point at the same time. 

 This problem is NP-hard since the area coverage problem even without consideration of energy constraint is proven to be NP-hard~\cite{arkin2000approximation}.

\begin{algorithm}
\caption{\textsc{AG-Cvg}}
\label{algorithm:ag_cover}
\textbf{Input:} binary map of environment $M$, \\
	\hspace*{1.20cm} $M_g$ and $M_a$ subregions\\
	\hspace*{1.20cm} $T$ battery capacity\\
	\hspace*{1.20cm} sensor footprint parameters $s_a$ and $s_g$\\
	\textbf{Output:} ${\pi}_a$ and ${\pi}_g$ paths
 \hrule
	\begin{algorithmic}[1]

 \State $\textit{$\pi_{a'}$} \gets \textrm{SolveEfficientBCDCoverage}(M_a, s_a)$
\State $\textit{$\pi_{g}$} \gets \textrm{SolveEfficientBCDCoverage}(M_g, s_g)$

\State $\textit{$G_{ag}$} \gets \textrm{BuildBipartiteGraph}(V(\pi_{a'}), V(\pi_{g}))$
\State $\textit{$match_{ag}[...]$} \gets \textrm{PerfectMatching}(G_{ag})$
\State $\textit{$t_{max}$} \gets \textrm{getMaxRVCost}(match_{ag}[...])$


\State $\textit{${Vc}_a$} \gets \textrm{ClusterPaths}(\pi_{a'}, T-t_{max},\nu_a)$
\State $\textit{${Vc}_g$} \gets \textrm{ClusterPaths}(\pi_{g}, T, \nu_g)$

\State $\textit{$Gc_{ag}$} \gets \textrm{BuildBipartiteGraph}(Vc_a, Vc_g)$ 
\State $\textit{$g_{match}[...]$} \gets \textrm{PerfectMatching}(Gc_{ag})$

\State $\textit{$i$} \gets 0$
\For{\textbf{each} $vc_a \in {g_{match}}$}
\State $\textit{$p$} \gets \textrm{AddMatchingUGVVertex}(vc_a~path, i)$
\State $\textit{$i$} \gets i++$
\State append $p$ to $\pi_a$
\EndFor

\State \Return $\pi_a, \pi_g$
\end{algorithmic}
\end{algorithm}
\section{Proposed Approach}
\label{section:proposed_approach}

In this section, we propose a heuristic method to solve Problem~\ref{main_problem}. The method \textsc{AG-Cvg}, outlined in Algorithm~\ref{algorithm:ag_cover} and visually depicted in Fig.~\ref{fig:pipeline}, is based on boustrophedon decomposition coverage planning~\cite{xu2014efficient} and graph matching~\cite{skiena1998algorithm}. The main intuition behind this approach is that the UAV has to recharge after each $T$ units of time. As such we want to ensure that the coverage trajectory contains a rendezvous after every $T$ units of time. 
\invis{the general idea is to split the environment into clusters in a way that when UAV and UGV are finishing a coverage of a single cluster they must rendezvous. This allows us to reduce the problem of finding UAV-UGV rendezvous locations that minimize UAV travel and minimize UGV travel costs.}
In practice, UGVs are restricted in the locations they can traverse whereas UAVs can usually fly directly between any pair of points. Hence in this work, we propose that only one of the robots (UAV) needs to take the rendezvous detour, and the other (UGV) will simply perform its own coverage task and rendezvous on its own coverage path. 

\begin{figure*}[h!]
      \centering
      \includegraphics[width=0.92\textwidth]{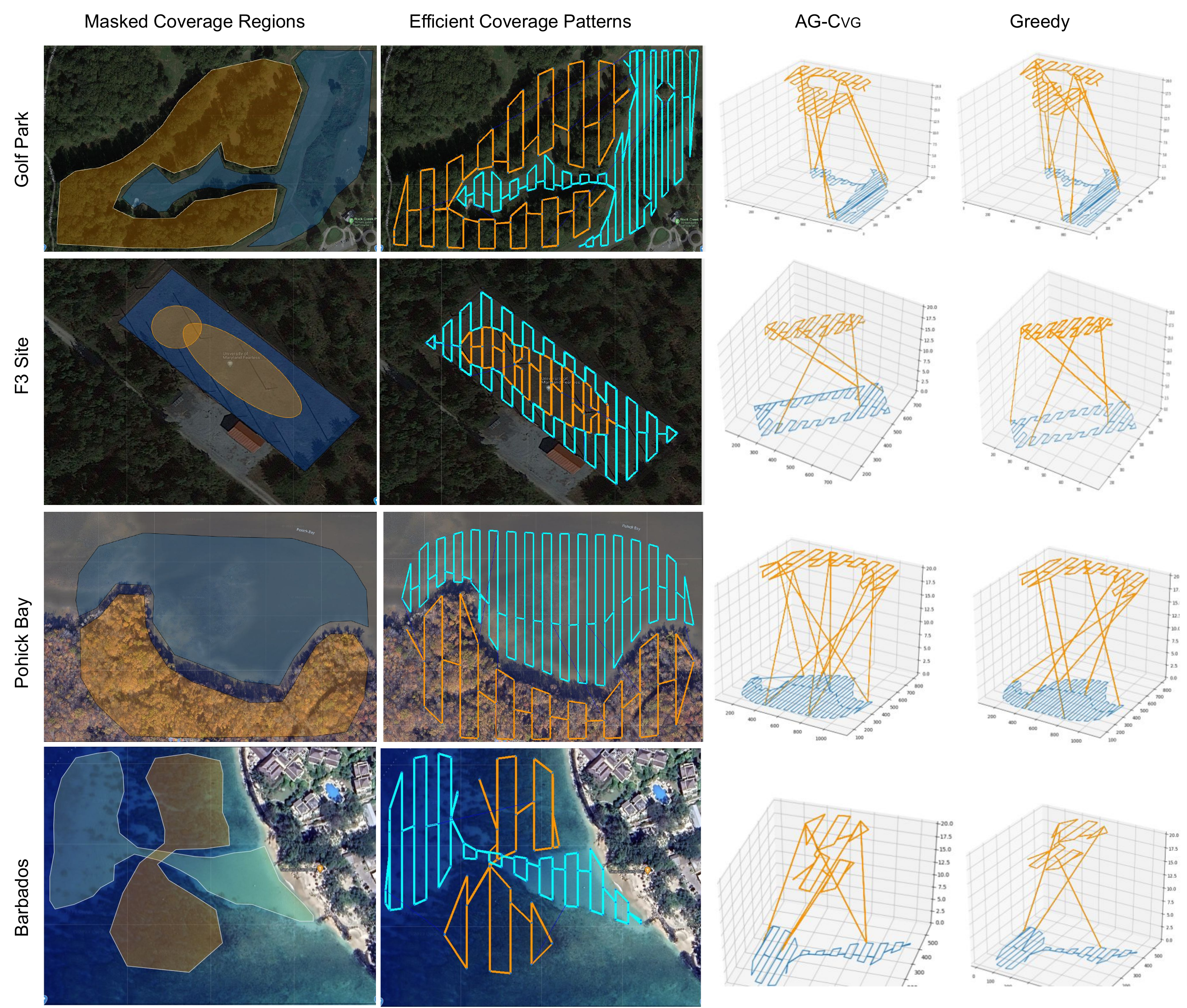}
     \caption{Samples of different environments and different scenarios of monitoring operations required from each robot. Light blue regions denote the regions for vehicles that can serve as mobile recharging stations (UGV, ASV), and the light orange one for vehicles with limited energy capacity (UAV). The Greedy method takes the returns back to the same location as it has started the rendezvous, whereas \textsc{AG-Cvg} due to perfect matching might often return to a different cluster to perform the coverage --- minimizing total rendezvous cost per survey mission.}
      \label{fig:sims}
\end{figure*}

The \textsc{AG-Cvg} algorithm comprises four main steps (see Algorithm~\ref{algorithm:ag_cover}). The first step (lines 1-2) (Fig.~\ref{fig:pipeline}A) generates complete coverage trajectories ${\pi_{a'}, \pi_{g}}$ using an efficient single robot coverage algorithm based on BCD~\cite{xu2014efficient} for both the UGV and UAV. This produces an ordered list of waypoints for each vehicle. 

The second step (lines 3--5) (Fig.~\ref{fig:pipeline}B) involves estimating the maximum rendezvous cost by 
building a bipartite graph with vertices $V(\pi_{a'}), V(\pi_{g})$ and edges connected between them. Note that the edge cost --- distance between UAV and UGV waypoints --- is calculated in the $xy$-plane, since UAV altitude is constant during the coverage. The algorithm then finds the maximum matching cost $t_{max}$ solving a maximum flow algorithm~\cite{skiena1998algorithm}.

 The third step (line 6--7) (Fig.~\ref{fig:pipeline}C), clusters the coverage trajectories such that a segment of the path in each cluster can be traversed by the UAV within the available battery capacity $T$, taking into account worst-case rendezvous time, i.e., the length of each segment is $(T-2t_{max})\nu_a$. For the UGV, since it does not need to leave its path for a rendezvous, the length of the path segment within a cluster is $T\nu_g$. The resulting clusters for the UAV and the UGV are denoted as $Vc_a$ and $Vc_g$ respectively. 

 \begin{figure}[h!]
      \centering
      \includegraphics[width=0.49\textwidth]{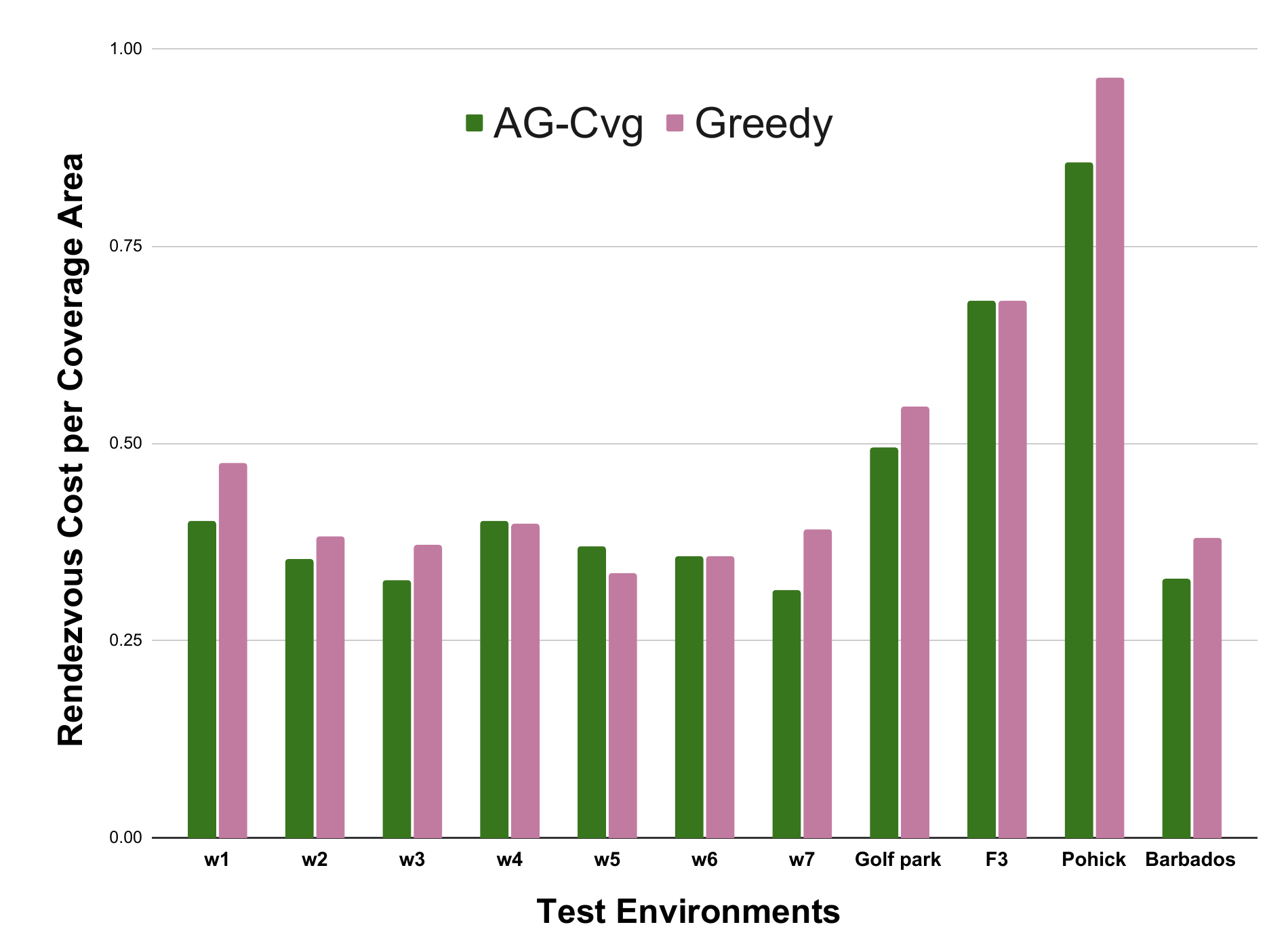}
      \vspace{-3mm}
     \caption{The rendezvous overhead is the same on the F3 site, whereas it is lower on other scenarios with 11.33$\%$ improvement on average, in some instances reaching up to 25$\%$ improvement.}
     \label{fig:result}
\end{figure}

 Finally step four (lines 8--16)~(Fig.~\ref{fig:pipeline}D), involves finding the rendezvous locations by building another bipartite graph between the clustered paths of the UGV and UAV and finding the perfect matching between them. The edge costs between the vertices of $Vc_a$ and $Vc_g$ are the time taken by the UAV to travel from the end/beginning of the cluster in $Vc_a$ to the end/beginning of the cluster in $Vc_g$. This matching ensures that all the segments of the coverage path are traversed and none of them is repeated. If the path segment $vc_a\in Vc_a$ is matched with the segment $vc_g \in Vc_g$, the UAV goes for a recharging rendezvous to the endpoint of $vc_g$ after traversing the path segment $vc_a$. The UAV traverses one of the segments, say $vc_a \in Vc_a$, and then goes for rendezvous to the endpoint of the UGV's segment (lines 11--16). In addition, depending on the size of UAV and UGV trajectory clusters $|{Vc_a}|$ and $|{Vc_g}|$, the UAV path $\pi_a$ will be defined in the following way:
    \subsubsection{If $|{Vc_a}| = |{Vc_g}|$} the UAV path will consist of all the vertices contained in the original coverage path where the order corresponds to UGV's path order in ${\pi_g}$. After completing a rendezvous, the UGV moves to its next segment, say $vc_g'$, and the UAV moves to the segment that is matched with $vc_g'$.
    \subsubsection{If $|{Vc_a}| > |{Vc_g}|$} the UGV will act as a static recharging location at the end of $\pi_g$. The UAV will rendezvous normally for the segments that have a bipartite matching with UGV segments, whereas, for the rest of the segments it will rendezvous with the UGV at the end of the UGV's path.
    \subsubsection{If $|{Vc_a}| < |{Vc_g}|$},the UAV trajectory will have idle periods when it is carried from UGV's one location to the other until UGV reaches a cluster that has a matching UAV cluster. For the rest of the matched UAV clusters, the path is created the same way as in the first step.

\invis{
\For{\textbf{each} $\textit{$m_a, m_g$} \in \textit{$M_a, M_g$}$}
\State $\textit{$G_a$} \gets \textrm{AddEdgeBetweenRegions}(m_a, m_g)$
\EndFor
\State $\textit{$\pi(m_a), \pi(m_g)$} \gets \textrm{TSPSolver}(G_a, G_g)$
}

\section{Experiments}
\label{section:experiments}

We performed extensive testing on maps of different sizes and different regions as assigned areas for coverage. Our experimental setup is designed to show the execution of the complete system - from generating paths on satellite images to real-world waypoint execution.

\subsection{Simulation Results}
\label{subsection:simulation_results}

We compare our method \textsc{AG-Cvg} with a greedy approach, where instead of solving the matching problem (Algorithm~\ref{algorithm:ag_cover} line 9), to find rendezvous UAV chooses to rendezvous at the closest UGV cluster/path segment. The experimental setup consists of 11 different coverage scenarios with different layouts and scales. Four scenarios are depicted in Fig.~\ref{fig:sims} to illustrate the resulting missions. The input consists of the satellite images of task areas with predefined masks indicating the regions assigned to UAV and to UGV. It should be noted that the proposed method is an offline approach and does not make any assumptions about the type of vehicle. Therefore, this method can be applied to any cooperative coverage task where the robots are capable of performing waypoint navigation. \invis{As such the selected three scenarios, we consider two experimental scenarios: the first one is about monitoring with a team of UAV and UGV, and the other uses a team of UAV and Autonomous Surface Vehicle (ASV).} 

Since the base algorithm for coverage is an efficient complete coverage method~\cite{xu2014efficient}, we compare the rendezvous overhead as a fraction of the original coverage trajectory that the UAV would have taken (Figure~\ref{fig:result}), e.g. $\frac{(L_1-L_2)}{L_1}$, where $L_1$ denotes the actual length of the coverage trajectory and $L_2$ denotes the length of the coverage trajectory without the energy constraint. The results show that our method -- \textsc{AG-Cvg} outperforms on average by at least 11.33\% greedy method and in some scenarios even reaches 25\% improvement. \invis{To put this in perspective, this means saving on average one-hour operational time on applications of: (1) 9 1 hour-long missions of search and rescue operations; (2) 9 hour military surveillance mission; (3) 15 coverage tasks running in parallel on a smart harvesting farm within one hour.}

\invis{
\begin{table*}
  \centering
\caption{The Simulation Results}\label{tab:res}  
\begin{tabular}{| c | c | c | c|c |}
   \hline
   & Greedy Method & PM Method & 2way PM Method & Best + CMDP \\
   \hline
   Total Travel of Rendezvous ($\%$) & 43.3 $\%$&	41.4$\%$ &	8.9$\%$ & 16.17$\%$ \\
  \hline
  UGV Total Coverage Cost w-o rendezvous & 86.75px & 46px & NA & NA \\ 
   \hline
   UAV Total Coverage Cost w-o rendezvous & 86.75px & 46px & NA & NA \\ 
   \hline
  UGV Total Coverage Cost with Rendezvous & 137.5 & 50 & NA & NA \\
   \hline
    UGV Total Coverage Cost with Rendezvous & 137.5 & 50 & NA & NA \\
   \hline
   UAV wait time overhead &. &.&.& .\\ \hline
   UGV wait time overhead &. &.&.& .\\ \hline
   Execution Time & 29.39\% & 31.05\% & 92.65\% & 91.42$\%$\\
   \hline
   Success Rate with stochastic energy & 29.39\% & 31.05\% & 92.65\% & 91.42$\%$\\
   \hline
  \end{tabular}
\end{table*}
}
\begin{figure*}[h!]
      \centering
      \includegraphics[width=1\textwidth]{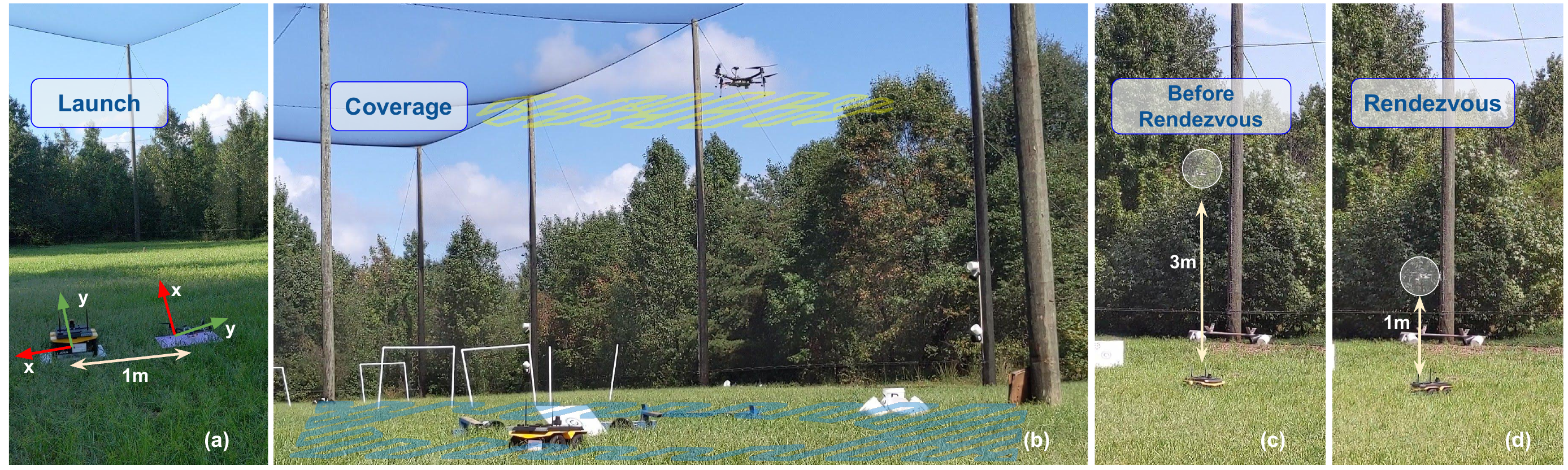}
     \caption{Proof of concept field trials with VOXL m500 UAV and Clearpath Jackal UGV at Fearless Flight facility (F3) at University of Maryland, College Park, MD, USA. (a) The vehicles are launched from a predefined location with $1m$ offset between their origins. (b) UAV and UGV are performing coverage tasks until they reach the rendezvous location. (c) Before rendezvous each of the vehicles waits for the other one to arrive at the location, UAV is at the flight altitude $3m$. (d) If both arrive UAV will rendezvous by descending to $1m$ altitude.}
      \label{fig:f3_2}
\end{figure*}

\begin{figure*}
      \centering
      \includegraphics[width=1\textwidth]{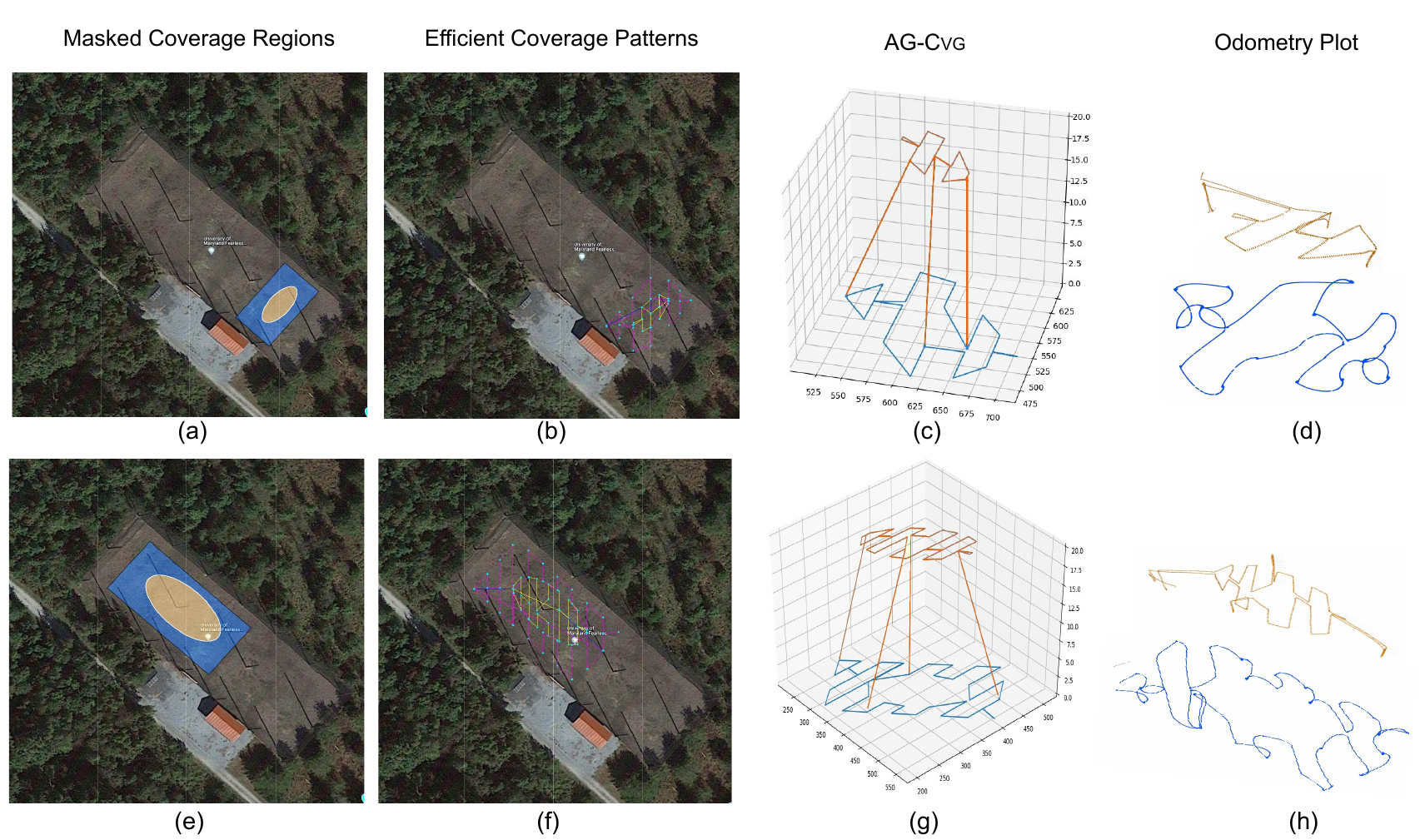}
     \caption{Execution of the coverage trajectories on a VOXL m500 drone and Clearpath Jackal ground vehicle at Fearless Flight facility (F3) at University of Maryland, College Park, MD, USA. The yellow trajectory is the UAV's coverage path and the purple one UGV's. The top row corresponds to the experiment on a small region with a size of $6.4m$ x $10m$, while the bottom row is the larger region with $10m$ x $25m$. The resulting trajectory plots of the robots' odometry readings are depicted in the rightmost column. Note, the UAV does not land on rendezvous but instead descends to $1m$ altitude.}
      \label{fig:f3_1}
\end{figure*}

\subsection{Field Deployment}

\subsubsection{UGV Platform}
We use Clearpath Jackal UGV equipped with
Hokuyo UST-10LX Lidar and IMU for localization. The onboard computer is Jetson AGX Xavier running Ubuntu 20.04 with ROS Noetic. The maximum speed of the vehicle is 2m/s but we set it to 1m/s in all experiments. The generated waypoints are relative to UGV's start position and the coordinate frame follows the right-hand rule: $Y$ aligned with the front of the robot, $X$ - left side (see Fig.~\ref{fig:f3_2}a)

\subsubsection{UAV Platform}
We used VOXL m500 UAV equipped with PX4 controller, ModelAI OV7251 Stereo camera for localization, and range-sensor for altitude estimation. The onboard computer is the VOXL flight board installed with Linux-based voxl operating system\invis{that runs a docker with Ubuntu 20.04 and ROS Noetic}. The total operational time in the air is 20 minutes. The maximum speed of the platform has been set to 1 m/s. The PX4 SDK was used to interface with the UAV and ROS. The generated waypoints are relative to UGV's frame with (0,1) starting point. The UAVs coordinate frame is according to the left-hand rule: $X$ aligned with the front of the drone, $Y$-right side, and $Z$ aligned with altitude (Fig.~\ref{fig:f3_2}a). 

\subsubsection{Execution}
The experiments were conducted at the Fearless Flight Facility (F3) at the University of Maryland, College Park (Fig.~\ref{fig:f3_2} - ~\ref{fig:f3_1}). F3 is a secure facility for outdoor drone tests with 100-foot wide, 300-foot long, and 50-foot high dimensions. We performed two flights with different sensor footprints and battery capacity values on a small region with a size of $6.4m$ x $10m$ and a larger region with $10m$ x $25m$ size (see Fig.~\ref{fig:f3_1}). In the smaller region, we set the battery capacity 2x smaller than in the larger environment to generate at least three rendezvous points, while in the larger region, we have 4 rendezvous locations. 

We use Ground Station (GS) equipped with Intel Core i9, running Ubuntu 18.04 and ROS Melodic. Both UAV and UGV use the GS as the ROS Master for the communication through AX21 Wi-Fi router. The \textsc{AG-Cvg} algorithm is implemented as a C++ library, that also provides flight-ready mission plans with respect to the selected launch position (Fig.~\ref{fig:f3_2} (a)).\invis{It provides methods to produce as an output the converted coordinates in the corresponding coordinate frame with respect to the selected launch position (Fig.~\ref{fig:f3_2} (a)).}The resulting mission files are passed as an input to the ROS-based packages that handle the rendezvous mission. While the mission point is not a rendezvous location both vehicles perform the coverage with UAV flying on a constant altitude of 3m (Fig.~\ref{fig:f3_2} (b)). Once the rendezvous location has been reached each of the vehicles is waiting for the other vehicle to arrive (Fig.~\ref{fig:f3_2} (c)). When rendezvousing, the UAV descends to 1m altitude and hovers (Fig.~\ref{fig:f3_2} (d)) (we do not address precision landing in this paper).
\section{CONCLUSIONS}
\label{section:conclusions}

In conclusion, this paper presents a comprehensive approach to solving the problem of coverage path planning for energy-constrained UAV and UGV teams. Our approach is motivated by the need to minimize rendezvous time while ensuring successful coverage. We proposed a heuristic method to solve this problem by clustering segments of the coverage
paths before using graph matching to find rendezvous for recharging. We conducted numerical analysis on real-world monitoring applications showing that our approach outperforms the Greedy approach, reducing rendezvous overhead by an average of 11.33\%, with up to 25\% improvement in some cases. We also successfully deployed our approach on a team consisting of a VOXL m500 drone and Clearpath Jackal ground vehicle - showing the full system execution from the satellite map to the real world. Our results demonstrate the feasibility and effectiveness of our approach in addressing the coverage path planning problem for energy-constrained UAV and mobile recharging UGV teams.

\textbf{Limitations and Future Work:}
The presented approach AGCvg is an offline coverage and rendezvous method that assumes that the energy consumption is deterministic. This method gives us a strategy on how to perform coverage when the energy is drawn at a constant rate, though when operating in the real world this usually will not be the case. To handle stochastic energy consumption, applying risk-aware online decision-making methods such as those proposed by Shi~et~al.~\cite{shi2022risk} to different clusters is an interesting direction for future work.




\addtolength{\textheight}{-4cm}
\bibliographystyle{IEEEtran}
\bibliography{IEEEabrv,root}

\end{document}